\documentclass{article}
\usepackage{ijcai23}

\usepackage{times}
\usepackage{soul}
\usepackage{url}
\usepackage{multirow}
\usepackage[hidelinks]{hyperref}
\usepackage[utf8]{inputenc}
\usepackage[small]{caption}
\usepackage{graphicx}
\usepackage{amsmath}
\usepackage{amsthm}
\usepackage{booktabs}
\usepackage{algorithm}
\usepackage{algorithmic}
\usepackage[switch]{lineno}

\title{NIPD: A Federated Learning Person Detection Benchmark Based on Real-World Non-IID Data}

\author{
Kangning Yin$^{1,2}$
\and
Zhen Ding$^1$\and
Zhihua Dong$^1$\and
Dongsheng Chen$^1$\and
Jie Fu$^1$\and
Xinhui Ji$^2$\and
Guangqiang Yin$^{1, *}$\and
Zhiguo Wang$^{1, *}$
\affiliations
$^1$School of Information and Software Engineering, University of Electronic Science and Technology of China, China.\\
$^2$Institute of Public Security, Kash Institute of Electronics and Information Industry, China.\\
\emails
knyin@std.uestc.edu.cn,
zhending@std.uestc.edu.cn,
zhdong@std.uestc.edu.cn,
chendongsheng@std.uestc.edu.cn,
fujie@std.uestc.edu.cn,
xinhj9981@163.com,
yingq@uestc.edu.cn,
zgwang@uestc.edu.cn.
}

\begin{document}

\maketitle

\begin{abstract}
Federated learning (FL), a privacy-preserving distributed machine learning, has been rapidly applied in wireless communication networks. FL enables Internet of Things (IoT) clients to obtain well-trained models while preventing privacy leakage. Person detection can be deployed on edge devices with limited computing power if combined with FL to process the video data directly at the edge. However, due to the different hardware and deployment scenarios of different cameras, the data collected by the camera present non-independent and identically distributed (non-IID), and the global model derived from FL aggregation is less effective. Meanwhile, existing research lacks public data set for real-world FL object detection, which is not conducive to studying the non-IID problem on IoT cameras. Therefore, we open source a non-IID IoT person detection (NIPD) data set, which is collected from five different cameras. To our knowledge, this is the first true device-based non-IID person detection data set. Based on this data set, we explain how to establish a FL experimental platform and provide a benchmark for non-IID person detection. NIPD\footnote{https://github.com/ShenSuanZiZhen/NIID\_Person\_Detection} is expected to promote the application of FL and the security of smart city. 
\end{abstract}

\section{Introduction}
As one of the big data analysis methods, machine learning (ML) has surpassed human performance in many fields, such as communication, security, and industrial manufacture. In wireless communications and networks, it is a forward-looking concept to combine Internet of Things (IoT) devices with artificial intelligence to form Artificial Intelligence Internet of Things (AIoT). AIoT can collect a large amount of data, and IoT is data-driven and intelligently connected through the analysis and processing of data. However, while IoT devices are capable of collecting large amounts of data, strict data privacy acts allow for stricter protection of data in the IoT and form isolated islands of data from device to device. The inaccessibility of private data and the vast communication overhead require to transfer the raw data to a central ML processor, which is difficult to apply traditional ML algorithms in wireless communications \cite{niknam2020federated}.

To address the issues of feature distribution and data privacy, federated learning (FL) \cite{mcmahan2017communication} has been proposed. FL is a distributed collaborative computing framework, as shown in Fig. 1. This framework provides a valuable solution for implementing artificial intelligence algorithms on these devices and enabling network edge intelligence in future sixth generation (6G) systems \cite{zhao2021federated}. The framework keeps data on edge devices that can locally implement image acquisition and training tasks, then transfers training parameters to the central server for aggregation. Replacing raw data transfer with shared model parameters significantly reduces communication overhead.

With the rapid development of deep learning and computing power, person detection has been widely used in real-world security. Person detection mainly relies on a large number of person image data to learn the characteristics of people. Due to the single environment of data collected by a single monitoring device, the trained model effect is not necessarily available in other real-world scenarios. FL can help train models to effectively adapt to changes in these systems while maintaining user privacy \cite{li2020federated}. Therefore, there has been a general trend to introduce FL in computer vision for IoT. Usually, the underlying assumption in FL is based on independent and identically distributed (IID) data. However, the assumption is difficult to hold in the real world. Data in the real world is often collected from different devices, which are not uniformly distributed. Due to the highly dynamic environment and person behavior in wireless communication networks, the collected data is non-IID. The minimum empirical error model obtained on the non-IID training set does not necessarily perform well on the testing set.

Moreover, there is a lack of real available data sets in the IoT, and most of the existing experiments \cite{li2022federated} for non-IID data are based on MNIST and CIFAR-10. A non-IID classification data set named NICO, containing animals and vehicles, is provided in \cite{he2021towards}, but is not practical in the computer vision domain of IoT. The object detection data sets Street\_5 and Street\_20 \cite{luo2019real} are somewhat segmented by device, but the amount of data is too small to validate the non-IID pervasiveness laws well.  \cite{chiu2020semisupervised,amit2022federated} conducted studies related to non-IID object detection, but did not open their data sets for other researchers to study. Therefore, data sets that can be applied for real-world IoT FL are urgently needed.

In this paper, we propose a non-IID person detection data set: non-IID IoT person detection (NIPD).  NIPD is generated from multiple cameras, and the labels of the data are first automatically filtered and labeled by an algorithm, and then manually filled in and accurately labeled. We evaluate the person distribution in the data set and there was significant variability in the distribution of data under each camera. Based on the FedVision \cite{liu2020fedvision}, we validate this data set using two classical two-stage and one-stage object detection algorithms, Faster R-CNN \cite{ren2015faster} and YOLOv3 \cite{redmon2018yolov3}. The results show that our NIPD is suitable for the IoT FL, providing a benchmark for non-IID person detection research in the monitoring field. The data set will facilitate research on non-IID IoT FL issues and promote computer vision security for smart city.

\begin{figure}[!t]
	\centering
	\includegraphics[width=3.5in]{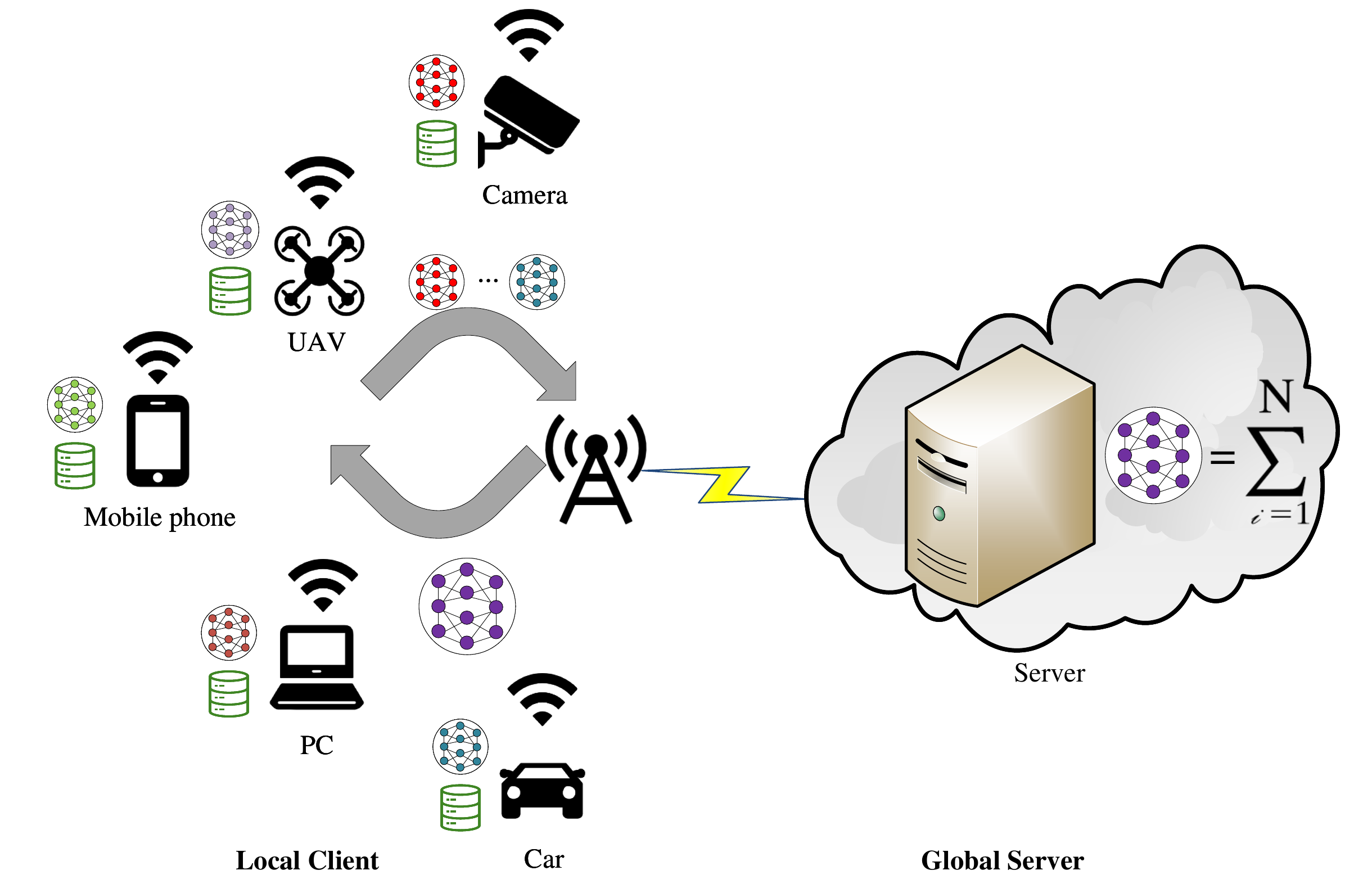}
	\caption{Federated learning in the wireless network IoT.}
	\label{fig_1}
\end{figure}

The rest of the paper is structured as follows. In Section II, we describe the basic concepts behind the proposed experimental design, and in Section III, we detail the experimental platform's practical implementation and experimental specification. Section IV describes NIPD and provides two experimental setups. Finally, Section V concludes the paper.

\section{Experimental platform design}
In this section, we first introduce the classification and impact of non-IID data. Then, we focus on how to design and build an experimental platform for IoT FL with non-IID data.

\subsection{Classification and impact of non-independently identically distributed data}
In a real-world FL environment, the data between participants is characterized by a high degree of heterogeneity and large gaps in data volume. Therefore, it is possible that the data between participants in different contexts are entirely different. Suppose the data sample is $(x, y)$, where $x$ is the input attribute or feature and $y$ is the label. For non-IID data, the local data distribution of terminal device $k$ is assumed to be $p_{\text k}(x, y)$, $p_{\text k}(x, y)$ is different from that of other terminal devices. Non-IID data are classified in \cite{kairouz2021advances} into the following five types:

\subsubsection {Feature distribution skew-covariate shift}
The same feature is expressed differently by different clients. For example, the same number is written differently by different people. This is the most common problem faced in FL.

\subsubsection {Label distribution skew-prior probability shift}
The same label, whose distribution depends on the client. For example, kangaroos are only found in Australia or zoos, which can cause different clients to have different preferences for prediction models.

\subsubsection {Same label, different features-concept drift}
For different clients, the same label has different features. For example, buildings differ significantly from one region to another. We believe that this problem is common in the field of video surveillance.

\subsubsection {Same features, different label-concept shift}
The same feature vector in the training data can have different labels due to personal preferences. For example, the label reflecting the emotions or predicting the next word differs according to the individual and the region. This problem is complex and makes personalized FL a big challenge.

\subsubsection {Data skewing}
 Different clients can save different data, which can cause unfairness in FL.

In either case of non-IID, the data distribution in each device is not representative of the global data distribution. The aggregated model does not perform well enough or fairly enough for the participating clients. Our NIPD considers the three types of non-IID described above: feature distribution skew-covariate shift, same label but different features-concept drift and data skewing.

\subsection{Introduction to the IoT FL experiment platform for non-IID data}

The experiment platform is designed to include data annotation, device communication, client-side training, server aggregation of models and distribution of the updated models.

\subsubsection {Data annotation}
This process converts unprocessed speech, images, text, video and other data into machine-recognized information, which requires the staff to take the position of the object of interest (e.g., person, bicycle) in the given image. The finished data can be used as training and testing samples for the object detection algorithm. Meanwhile, new image data is obtained from the camera and staff continue to annotate it, which can update the training samples and  achieve online learning.

\subsubsection {Device communication}
 Communication between the server and the client takes place during the FL process. First, the server initiates a FL task to the clients, and the client participating in the training respond to the request of the server. The server then selects the clients to participate in the training, issues the initial model, and the clients train the model on the local data and upload it to the servers. Finally, the server aggregates the models and sends them to the clients for further training until convergence. Therefore, in the design of the experimental platform, the stability of communication and the cost of communication are taken into account.

\subsubsection {Client training}
The client is trained locally after receiving the FL task. Each client should have a configuration information file, including the number of local training epoch, batch size, optimizer, learning rate, the number of reconnections, etc. The training process is similar to traditional machine learning algorithms, where communication with the server is required, as well as updating the training model each round.

\subsubsection {Model aggregation}
After the clients have been trained on the local model, the model parameters for each client are transferred to the FL server. The server stores the updated model parameters. In terms of aggregation algorithms, there are several options for dealing with the adverse non-IID problems \cite{li2022federated}. In addition, security and privacy need to be considered, which prevents attack methods such as backdoor attacks and inference \cite{gao2022secure}.

\section{Experimental Platform Implementation}

In this section, we explain how to implement the design of the experimental platform for non-IID data in IoT devices described in the previous section. First, we describe the challenges of the implementation, including the client-side data annotation of the experimental platform and how it was communicated, the further partitioning of the partitioned existing data set with a quantitative imbalance, and the aggregation algorithm used. Then, we present the exact experimental specifications of the experiment and give examples of the experimental setting and its results.

\subsection{Challenges in the implementation process}

Non-IID data poses a huge challenge to the learning accuracy of FL algorithms. Since the data distribution of users in different regions and environments is non-IID, the models of users are also different. If they are directly considered as a class for uploading and aggregating, the generated models are bound to deviate from the target models. The existing FL algorithms cannot adapt to all cases. The experimental platform described in this paper addresses the non-IID data phenomenon prevalent in IoT and is designed in four aspects: data annotation, device communication, number imbalance division, and aggregation algorithms.

\subsubsection {Data annotation}
After the image data is captured by the IoT camera, it needs to be manually annotated before it can be sent to the FL client for learning. The data annotation software uses labelImg\footnote{https://github.com/HumanSignal/labelImg} and the annotation schematic is shown in Fig. 2. Manual image annotation is first performed, requiring accurate annotation of each object, which will result in a good global model of the effect. Then, after the client receives the task initiated by FL, the corresponding client reads the annotated data for training. Finally, after the training is completed, it is sent to the server side for model aggregation, model updating and opening the next round of training.

\begin{figure*}[!t]
	\centering
	\includegraphics[width=6.5in]{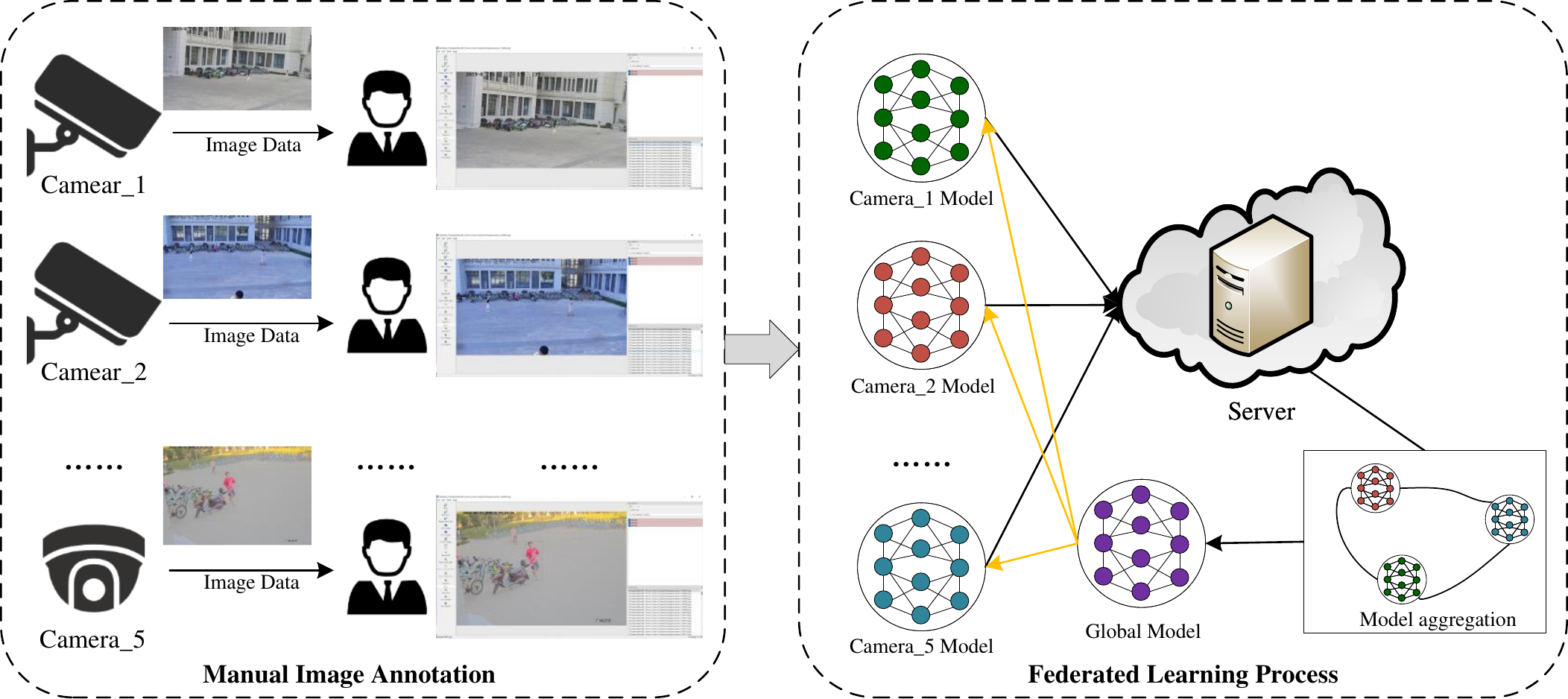}
	\caption{Data annotation process in wireless IoT FL.}
	\label{fig_sim}
\end{figure*}

\subsubsection {Device communication}
The communication part of the experiment platform is based on the Flask-SocketIO framework, and the communication process between the server and client is shown in Fig. 3. Firstly, each client waits for the server to initiate a federal learning request after collecting and manually annotating image data. Secondly, after the server initiates a training job request, the client makes a participating training response, and the server selects the clients to participate in this round of training. Thirdly, each participating client downloads the training request and receives the initial model from the server, while configuring the local training epoch, batch size, data set metrics, etc. for each client for local training. Fourthly, after the local training epoch requirement is met, the local model is uploaded to the server side. The server waits for all the participating clients' models to be uploaded in this round, then performs model aggregation and generates a global model. Finally, the server sends the global model to each client, and each client determines whether convergence is achieved based on the metrics measured on the test data.

\begin{figure}[!t]
	\centering
	\includegraphics[width=3.5in]{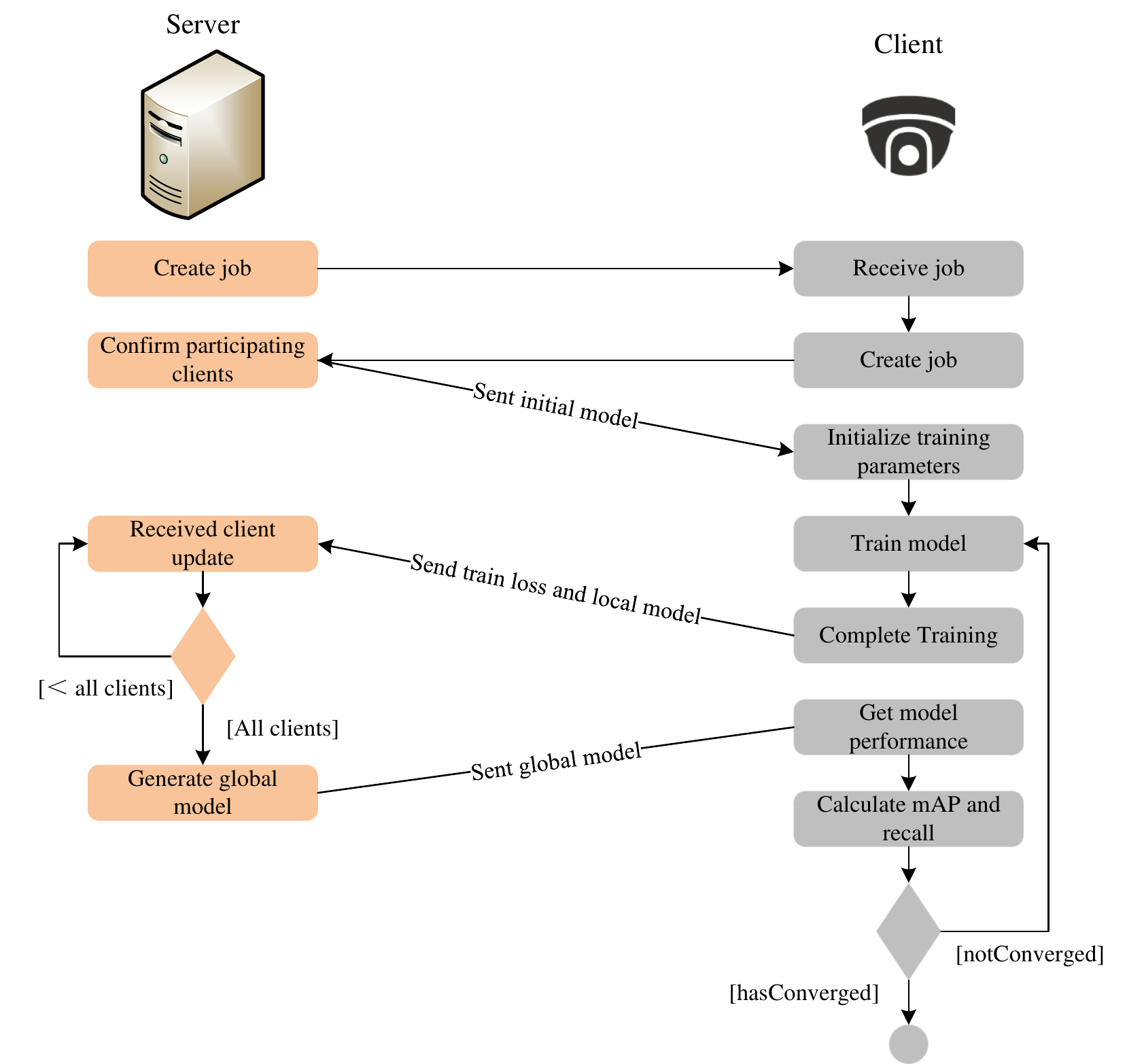}
	\caption{Communication process in wireless IoT FL.}
	\label{fig_1}
\end{figure}

\subsubsection {Quantity imbalance division}
 We design a hyperparameter ${\alpha}$ to further divide the data set so that the amount of data held under each client is imbalanced and more relevant to the real scene. ${\alpha}$ takes a range of 0 - 0.2, the larger it is, the more drastic the data imbalance is.

\subsubsection {Aggregation algorithm}
Our aggregation algorithm uses the classical aggregation algorithm Federated Averaging (FedAvg) \cite{mcmahan2017communication}. The server only passes the current new parameters to the selected model each time, aggregates the model parameters of the selected client, and selects the model aggregation weight according to the number of samples of the client.

Our person detection code is based on the architecture of FedVision \cite{liu2020fedvision}, and related experiments are performed on the Ubuntu 20.04 Linux operating system. The PyTorch version is 1.7.0, the CUDA version is 11.2, and the Python version is 3.8.3. The training is performed on a GPU server with two Intel Xeon Gold 6226R CPUs and eight Tesla T4 GPUs. The RAM of the server is 256 GB.
The experimental environment of our person detection algorithm is shown in Table 1.

\begin{table}
\caption{Experimental environment settings }
	\centering
	\begin{tabular}{ll}
		\toprule
		Settings & Details\\
		\midrule
		Servers    & Intel Xeon Gold 6226R CPU * 2       \\
	            	& Tesla T4 GPU * 8         \\
		Operating System  & Ubuntu 20.04       \\
		FL Framework &FedVision        \\
		Python & 3.8.3        \\
		Pytorch &1.7.0        \\
		CUDA & 11.2        \\
		\bottomrule
	\end{tabular}
	\label{tab:booktabs}
\end{table}

The number of participating clients can be fixed by the server, and the participating clients per round can be selected by configuring the value of the parameter NUM\_CLIENTS\_CONTACTED\_PER\_ROUND. The MAX\_NUM\_ROUNDS parameter is used to adjust the training rounds, with 50 rounds for default setting. We use two classic object detection networks: YOLOv3 and Faster R-CNN. We adopts Adam optimization method to train YOLOv3, the initial learning rate is 1e-3, the batch size of Camera\_1 to Camera\_4 is set to 6, and the batch size of Camera\_5 is set to 10. Faster R-CNN is trained using the Adam optimization method. The fixed learning rate is 1e-4, the momentum is 0.9, and the batch size from Camera\_1 to Camera\_5 is set to 1. It should be noted that the pre-trained model is loaded to speed up the convergence of the model, where the pre-trained Darknet-53 model is used for YOLOv3 and the pre-trained VGG16. In particular, in the data preprocessing part, the image input of the YOLOv3 algorithm is uniformly scaled to 416 * 416.

The local epoch of each client is set to 1 by default for both the Faster R-CNN and YOLOv3 object detection models.

The evaluation criterion we adopted is the general index mean Average Precision (mAP) in the field of object detection, and the mean value of the Average Precision (AP) value of each class is calculated. Intersection Over Union (IoU) represents the area of overlap between the predicted bounding box and the ground truth bounding box to determine whether the prediction is true positive.The specific formulas are as follows:
\begin{align}
mAP = \frac{{\sum\nolimits_{k = 1}^k {A{P_i}} }}{k}
\end{align}

\begin{align}
AP = \int_0^q {p\left( r \right)} dr
\end{align}

\begin{align}
IoU = \frac{{area\left( {{B_{gt}} \cap {B_{pred}}} \right)}}{{area\left( {{B_{gt}} \cup {B_{pred}}} \right)}}
\end{align}

where AP is calculated for each class separately and k is the number of classes (in our dataset, k is 1). And the IoU threshold is set to 0.5 to calculate the mAP.

In the example program, we perform the following two setups to give concrete examples of the experiments.

{\bf{Setting 1: }}Standard Non-IID data. The amount of data held by each client is basically the same, around 1600. From Camera\_1 to Camera\_5, the number of training data is 1590, 1574, 1581, 1583 and 1607, and the number of test data is 410, 426, 419, 417 and 393.

{\bf{Setting 2: }}The number of data is imbalanced, as reflected by the fact that there is a certain difference in the number of images under each camera. We set the size of the above hyperparameter $\alpha$ to 0.1, reclassifying the training data in Setting 1 as imbalanced. From Camera\_1 to Camera\_5, the number of training data is 1590, 1412, 1284, 1117, and 937, and the number of test data is the same as Setting 1.
\begin{figure}[!t]
	\centering
	\includegraphics[width=3.2in]{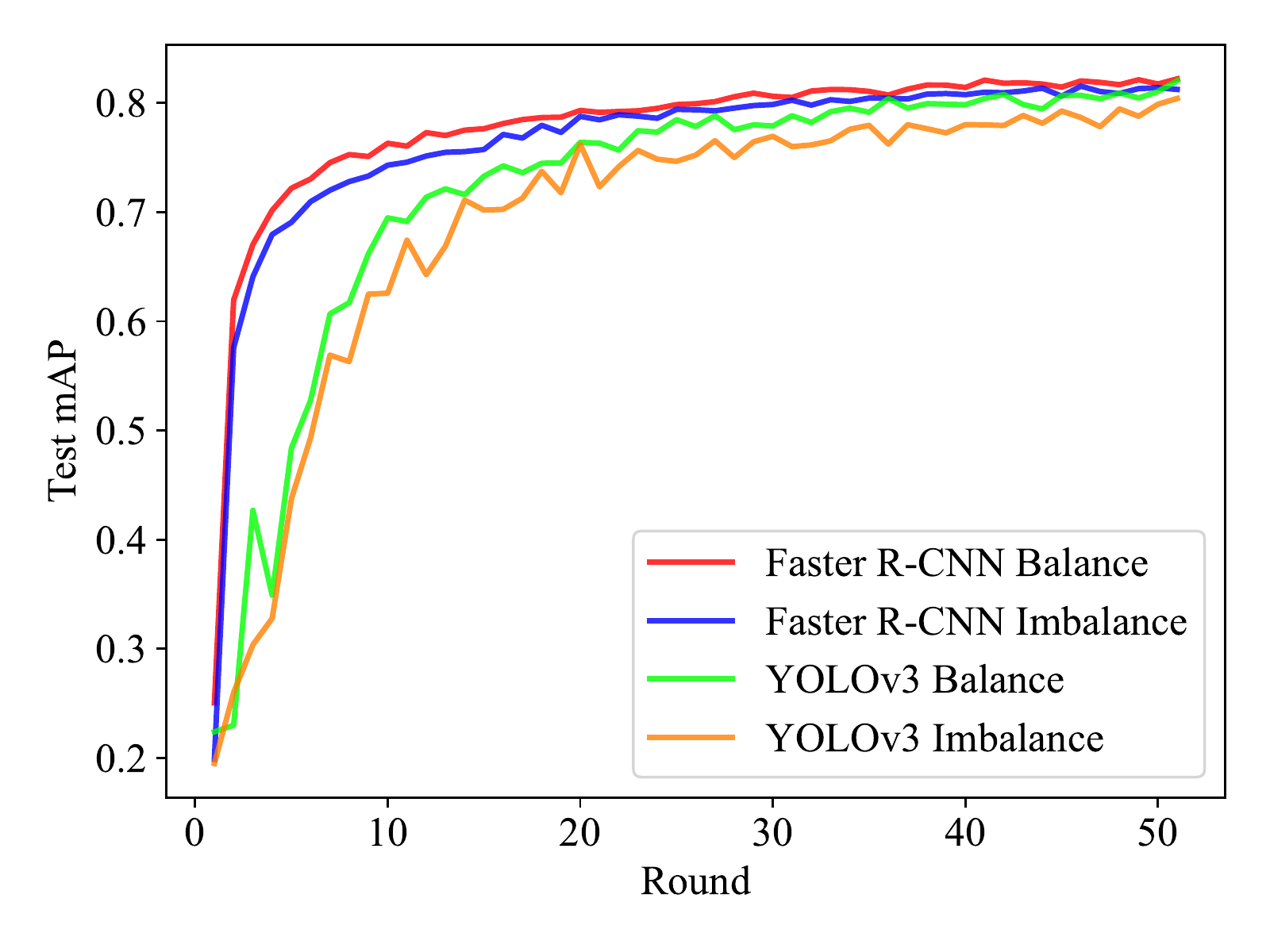}
	\caption{Person detection results for YOLOv3 and Faster R-CNN.}
	\label{fig_4}
\end{figure}

After completing the above configuration, we conduct our experiments. We use the methods Faster R-CNN and YOLOv3 to conduct 50 rounds of training for standard and imbalanced data sets respectively. The test mAP results are shown in Fig. 4. In the case of 50 rounds of training, the benchmark values we provide are as follows: The mAP of Faster R-CNN on the NIPD standard dataset is 82.2 percent, and the mAP at $\alpha$ = 0.1 is 81.5 percent. The mAP of YOLOv3 on the NIPD standard dataset is 82.0 percent, and the mAP at $\alpha$  = 0.1 is 80.4 percent.

\section{Data Set}

In this section, we provide an exhaustive introduction to the data set, with person sizes defined in the context of the scenario. The data clusters are introduced including camera locations, data annotation formats, object distribution, data partitioning, and some challenges in NIPD applications.

\subsection{Resource Description}

Current real-world non-IID data lack corresponding dedicated data sets in IoT-related devices. The data sets used by most researchers are base data sets for object detection domain data, but these data sets do not accurately model real device situations in IoT-related studies.

To solve this problem, we randomly capture different scenes from the cameras in the front plaza of the Innovation Centre of the University of Electronic Science and Technology of China (UESTC) at similar periods, and delete the pictures of people that do not exist within the field of view and the night pictures. We select the data of five representative cameras located around the square, named Camera\_1 to Camera\_5. These cameras' view angle, chromatic aberration, resolution, and mounting heights all differ, and the scene is shown in Fig. 5, which contains the camera resolutions and mounting heights. Camera\_1 to Camera\_4 have a resolution of 1920 * 1080 and Camera\_5 has a resolution of 3072 * 2048. Two thousand images are acquired from each camera, for a total of 10000 images, all of which are identified with a person class and ensure that at least one person exists within each image. The total number of all people in the data set is 55,160, with an average of 5.516 persons per image.

\begin{figure}[!t]
	\centering
	\includegraphics[width=3.4in]{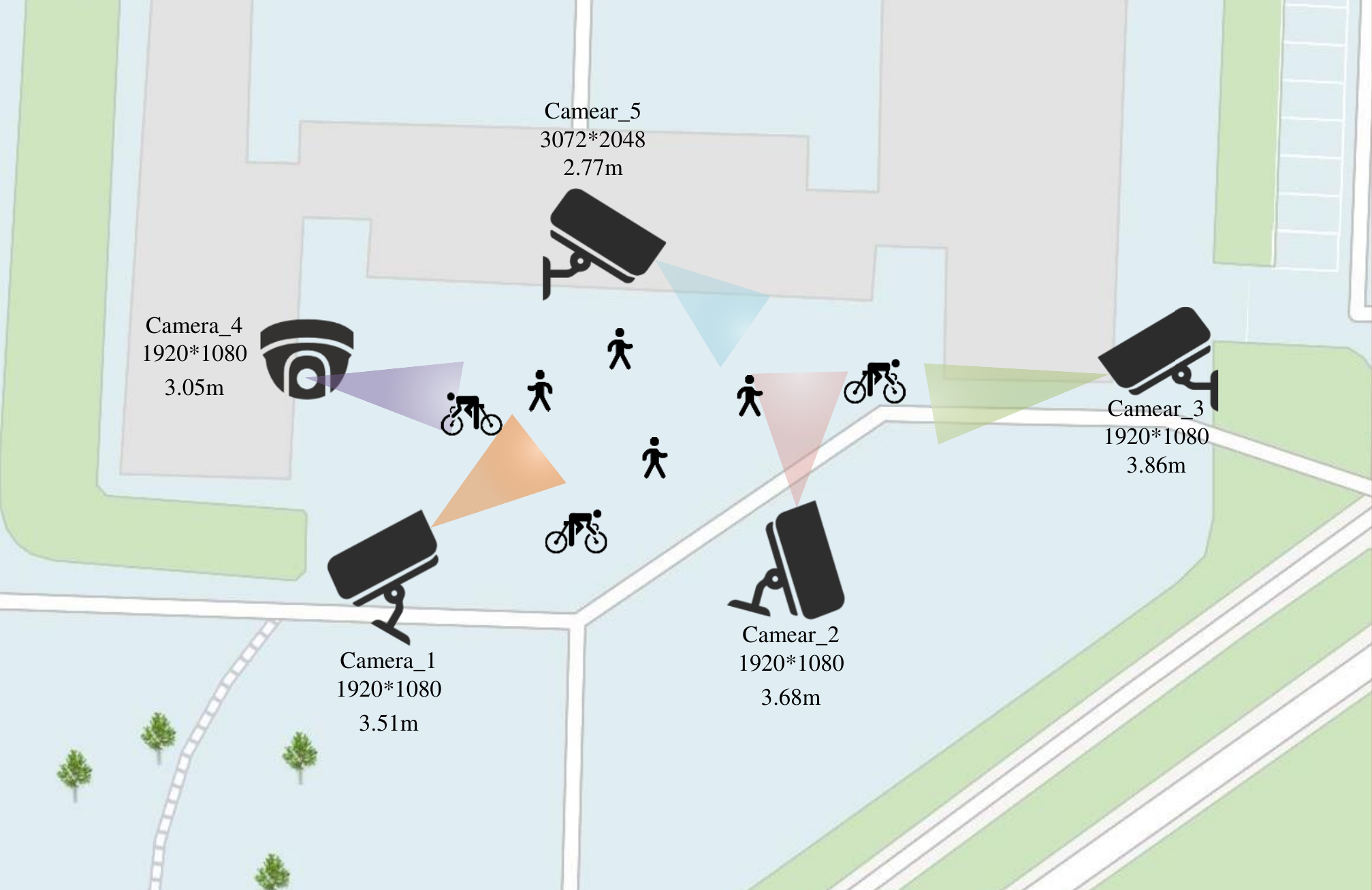}
	\caption{Camera data acquisition scene in UESTC Innovation Center.}
	\label{fig_1}
\end{figure}

\subsection{Data format}

All people are labeled using rectangular bounding boxes. To accommodate the need for pervasiveness of person detection, two types of labels are used to standardize the data set, the VOC format and the YOLO format. For the VOC format labeling, a line represents the person bounding box information in the form of
$\left\{ {x_{\text{min}} ,x_{\text{max}} ,y_{\text{min}} ,y_{\text{max}} } \right\}$, where the coordinates $x_{\text{min}},y_{\text{min}}$ are the top left corner of the bounding box and the coordinates $x_{\text{max}},y_{\text{max}}$ are the bottom right corner of the bounding box.
For the YOLO format of the label, each line describes the rectangular range information of the object in the form of $\left\{ {label,x,y,w,h} \right\}$,where label indicates the class of the object (in this data set only person), $(x, y)$ the center of the bounding box, $w$ the width of the bounding box and $h$ the height of the bounding box.

In the standard data set, we use approximately 80 percent of the data set (7,935 images) for training and the remaining approximately 20 percent (2,065 images) for testing. The number and size distribution of persons these images are also distributed with the images. The data set splits are all randomly selected to ensure the reliability and generalizability of the overall data set during use.

\begin{table*}[!t]
	    \caption{Sample statistics of NIPD data set\label{tab:table2}}
	\centering
	\begin{tabular}{ccrrrrr}
		\toprule
		{\bf{Type}} & {\bf{Camera}} & {\bf{Images}} & {\bf{Total objects}} & {\bf{Large objects}} & {\bf{Medium objects}} & {\bf{Small objects}} \\
		\hline
		\multirow{5}{*}{All} & Camera\_1 & 2000 & 10497 & 676 &  9497 & 324\\
		& Camera\_2 & 2000 & 7573 & 894 &  6417 & 262\\
		& Camera\_3 & 2000 & 12857 & 2059 &  9660 & 1138\\
		& Camera\_4 & 2000 & 13434 & 2453 &  8125 & 2856\\
		& Camera\_5 & 2000 & 10799 & 4430 &  6205 & 164\\
		\hline
		\multirow{5}{*}{Training}  & Camera\_1 & 1590 & 8303 & 548 & 7499 & 256\\
		& Camera\_2 & 1574 & 5979 & 698 & 5082 & 199\\
		& Camera\_3 & 1581 & 10121 & 1574 & 7656 & 891\\
		& Camera\_4 & 1583 & 10555 & 1950 & 6370 & 2235\\
		& Camera\_5 & 1607 & 8572 & 3493 & 4946 & 133\\
		\hline
		\multirow{5}{*}{Testing}   & Camera\_1 & 410 & 2194 & 128 & 1998 & 68\\
		& Camera\_2 & 426 & 1594 & 196 & 1335 & 63\\
		& Camera\_3 & 419 & 2736 & 485 & 2004 & 247\\
		& Camera\_4 & 417 & 2879 & 503 & 1755 & 621\\
		& Camera\_5 & 393 & 2227 & 937 & 1259 & 31\\
		\hline
		\multirow{5}{*}{{$\alpha = 0.1$}}   & Camera\_1 & 1590 & 8303 & 548 & 7499 & 256\\
		& Camera\_2 & 1411 & 5332 & 623 & 4527 & 182\\
		& Camera\_3 & 1284 & 8201 & 1302 & 6184 & 715\\
		& Camera\_4 & 1117 & 7430 & 1361 & 4479 & 1590\\
		& Camera\_5 & 937 & 4915 & 2050 & 2789 & 76\\
		
		\bottomrule
	\end{tabular}
\end{table*}

\subsection{Data division}
We aim to simulate real-world IoT FL with different regions and performance differences on surveillance cameras. This data partition is based on the real situation of the cameras and the data set suffers from non-IID data distribution problems. Combining the definition of object size in the Society of Photo-Optical Instrumentation Engineers (SPIE) and COCO data set, we take a relative size approach to the definition based on the characteristics of NIPD, where less than 0.12 percent of the image area is a small object, between 0.12 percent and 1.08 percent of the image area is a medium object, and greater than 1.08 percent of the image area is a large object.

Table 2 shows the detailed distribution of data and persons in different cameras according to the full data, the standard training data, the testing data, and the training data at $\alpha$ = 0.1. From this, we can see the bias of the data per client in this case, and the subsequent resulting bias in client training. In terms of the total number of persons, Camera\_4 has the highest number of total personss, 77.4 percent higher than Camera\_2, which has the lowest number of total persons. The different training number will lead to differences in the convergence degree of the model for different clients, and the training effect of clients with fewer persons will be weaker than those with more persons. In terms of large, medium and small persons, Camera\_5 has the highest number of large objects, which is 6.55 times higher than Camera\_1 with the lowest number of large persons; Camera\_3 has the highest number of medium persons, which is 1.56 times higher than Camera\_5 with the lowest number of medium persons; Camera\_4 has the highest number of small persons, which is 17.41 times higher than Camera\_5 with the lowest number of small persons. The networks trained by different clients have different scale preferences, which poses a great challenge for the aggregated object detection network to be able to identify persons at different scales simultaneously.
Thus, NIPD can effectively test the ability of IoT camera devices to solve real-world problems through FL, while serving as a benchmark to solve the non-IID data distribution problem in real-world applications.

\subsection{Challenges in NIPD applications}

We have reflected on non-IID FL in light of some meaningful observations in the proposed data set and suggest some possible guidelines for the design of subsequent experimental use.

\subsubsection {Object consistency}
 Due to different geographical locations, different devices have different object scale distributions, which may cause their local models to be biased towards the size of a particular object. Therefore, in subsequent experiments, the aggregation algorithm can be reconsidered in conjunction with the scale of objects in the model device to enhance the generalization capability of the final model.

\subsubsection {Selection of hyperparameter $\boldsymbol{\alpha}$}
To ensure imbalance in the amount of data, ${\alpha}$ can be selected from 0 - 0.2 in the subsequent experimental design according to the actual. The more severe the client data imbalance, the worse the aggregation model performs.

\subsubsection {Group Normalization}
 Batch Normalization (BN) is more likely to exacerbate the mismatch between the global mean and variance during non-IID training, thus affecting the validation accuracy. Therefore, \cite{hsieh2020non} proposed that Group Normalization (GN) can be used to replace the BN structure in the network, and GN can overcome the shortcomings of BN and Layer Normalization (LN) in the non-IID case to a certain extent, providing more accurate mean and variance. However, GN is not widely used at present, whose adoption in practical experiments is still debatable in the context of specific situations.

\subsubsection {Model Lightweighting}
 Since FL involves data exchange between multiple devices and model parameter update, communication efficiency is one of the important factors affecting the performance of FL. The model lightweight can reduce the parameter quantity and computational complexity of the model, and can reduce the communication bandwidth required by the device to transmit the model parameters, thereby improving the communication efficiency. For the YOLOv3 model, the number of model parameters can be reduced by using more lightweight network structure and model lightweight technologies such as pruning, thereby reducing communication cost. Taking Setting 1 as the experimental data set, we conducted some experiments, as shown in Table 3. Firstly, before training, the convolutional layer in YOLOv3 is replaced by depthwise seperable convolution (DS Conv). Compared with YOLOv3, the number of parameters is reduced by nearly half, and the number of parameters of the model uploaded and distributed by the client and server is reduced. When the model training meets certain mAP, the network slimming (NS) pruning \cite{liu2017learning}  is used to prune the server side. Under the condition of basically maintaining the mAP, the number of parameters is reduced by more than half again, which reduces the number of parameters issued by the server to the client.

\begin{table}
	\caption{Experimental environment settings }
	\centering
	\begin{tabular}{llrr}
		\toprule
		Number & Model & Number of  & mAP \\
		 &  &  parameters & \\
		\midrule
		0    & YOLOv3 & 61,523,734  & 82.0      \\
		1    & + DS Conv & 34,322,198  & 81.2      \\
		2  &+ DS Conv + NS & 16,342,727 &   80.9     \\
		\bottomrule
	\end{tabular}
	\label{tab:booktabs}
\end{table}

\section{CONCLUSION}

In this paper, we set up an IoT camera experimental platform for the lack of non-IID person detection data sets in the IoT field, presenting the initial concept to the practical implementation. We propose a non-IID person detection dataset named NIPD, which is the first data set proposed for the non-IID person detection problem with three non-IID types: feature distribution skew-covariate shift, same label but different features, and data skewing. Moreover, we have conducted experiments on this dataset using YOLOv3 and Faster R-CNN to provide a non-IID person detection benchmark for others to conduct subsequent studies. We hope to carry out the next work in two ways. On FL, we will improve the communication policy and aggregation algorithms to reduce the communication cost in the wireless communication, and generate more personalized models for devices. On the data set, we will subsequently improve on the data set by labeling more classes, such as bicycle, electric vehicle, and person attributes, to satisfy the class imbalance of the non-IID data set.

\section*{Acknowledgments}
This work was supported in part by the Natural Science Foundation of Xinjiang Uygur Autonomous Region  (No.2022D01B187).

\bibliographystyle{named}
\bibliography{ijcai23}

\begin{thebibliography}{}

\bibitem[\protect\citeauthoryear{Amit and Mohan}{2022}]{amit2022federated}
Rasna~A Amit and C~Krishna Mohan.
\newblock Federated learning: Dataset management for airport object
  representations using remote sensing images.
\newblock In {\em 2022 IEEE Aerospace Conference (AERO)}, pages 1--14. IEEE,
  2022.

\bibitem[\protect\citeauthoryear{Chiu \bgroup \em et al.\egroup
  }{2020}]{chiu2020semisupervised}
Te-Chuan Chiu, Yuan-Yao Shih, Ai-Chun Pang, Chieh-Sheng Wang, Wei Weng, and
  Chun-Ting Chou.
\newblock Semisupervised distributed learning with non-iid data for aiot
  service platform.
\newblock {\em IEEE Internet of Things Journal}, 7(10):9266--9277, 2020.

\bibitem[\protect\citeauthoryear{Gao \bgroup \em et al.\egroup
  }{2022}]{gao2022secure}
Jiqiang Gao, Baolei Zhang, Xiaojie Guo, Thar Baker, Min Li, and Zheli Liu.
\newblock Secure partial aggregation: Making federated learning more robust for
  industry 4.0 applications.
\newblock {\em IEEE Transactions on Industrial Informatics}, 2022.

\bibitem[\protect\citeauthoryear{He \bgroup \em et al.\egroup
  }{2021}]{he2021towards}
Yue He, Zheyan Shen, and Peng Cui.
\newblock Towards non-iid image classification: A dataset and baselines.
\newblock {\em Pattern Recognition}, 110:107383, 2021.

\bibitem[\protect\citeauthoryear{Hsieh \bgroup \em et al.\egroup
  }{2020}]{hsieh2020non}
Kevin Hsieh, Amar Phanishayee, Onur Mutlu, and Phillip Gibbons.
\newblock The non-iid data quagmire of decentralized machine learning.
\newblock In {\em International Conference on Machine Learning}, pages
  4387--4398. PMLR, 2020.

\bibitem[\protect\citeauthoryear{Kairouz \bgroup \em et al.\egroup
  }{2021}]{kairouz2021advances}
Peter Kairouz, H~Brendan McMahan, Brendan Avent, Aur{\'e}lien Bellet, Mehdi
  Bennis, Arjun~Nitin Bhagoji, Kallista Bonawitz, Zachary Charles, Graham
  Cormode, Rachel Cummings, et~al.
\newblock Advances and open problems in federated learning.
\newblock {\em Foundations and Trends{\textregistered} in Machine Learning},
  14(1--2):1--210, 2021.

\bibitem[\protect\citeauthoryear{Li \bgroup \em et al.\egroup
  }{2020}]{li2020federated}
Tian Li, Anit~Kumar Sahu, Ameet Talwalkar, and Virginia Smith.
\newblock Federated learning: Challenges, methods, and future directions.
\newblock {\em IEEE Signal Processing Magazine}, 37(3):50--60, 2020.

\bibitem[\protect\citeauthoryear{Li \bgroup \em et al.\egroup
  }{2022}]{li2022federated}
Qinbin Li, Yiqun Diao, Quan Chen, and Bingsheng He.
\newblock Federated learning on non-iid data silos: An experimental study.
\newblock In {\em 2022 IEEE 38th International Conference on Data Engineering
  (ICDE)}, pages 965--978. IEEE, 2022.

\bibitem[\protect\citeauthoryear{Liu \bgroup \em et al.\egroup
  }{2017}]{liu2017learning}
Zhuang Liu, Jianguo Li, Zhiqiang Shen, Gao Huang, Shoumeng Yan, and Changshui
  Zhang.
\newblock Learning efficient convolutional networks through network slimming.
\newblock In {\em Proceedings of the IEEE international conference on computer
  vision}, pages 2736--2744, 2017.

\bibitem[\protect\citeauthoryear{Liu \bgroup \em et al.\egroup
  }{2020}]{liu2020fedvision}
Yang Liu, Anbu Huang, Yun Luo, He~Huang, Youzhi Liu, Yuanyuan Chen, Lican Feng,
  Tianjian Chen, Han Yu, and Qiang Yang.
\newblock Fedvision: An online visual object detection platform powered by
  federated learning.
\newblock In {\em Proceedings of the AAAI Conference on Artificial
  Intelligence}, volume~34, pages 13172--13179, 2020.

\bibitem[\protect\citeauthoryear{Luo \bgroup \em et al.\egroup
  }{2019}]{luo2019real}
Jiahuan Luo, Xueyang Wu, Yun Luo, Anbu Huang, Yunfeng Huang, Yang Liu, and
  Qiang Yang.
\newblock Real-world image datasets for federated learning.
\newblock {\em arXiv preprint arXiv:1910.11089}, 2019.

\bibitem[\protect\citeauthoryear{McMahan \bgroup \em et al.\egroup
  }{2017}]{mcmahan2017communication}
Brendan McMahan, Eider Moore, Daniel Ramage, Seth Hampson, and Blaise~Aguera
  y~Arcas.
\newblock Communication-efficient learning of deep networks from decentralized
  data.
\newblock In {\em Artificial intelligence and statistics}, pages 1273--1282.
  PMLR, 2017.

\bibitem[\protect\citeauthoryear{Niknam \bgroup \em et al.\egroup
  }{2020}]{niknam2020federated}
Solmaz Niknam, Harpreet~S Dhillon, and Jeffrey~H Reed.
\newblock Federated learning for wireless communications: Motivation,
  opportunities, and challenges.
\newblock {\em IEEE Communications Magazine}, 58(6):46--51, 2020.

\bibitem[\protect\citeauthoryear{Redmon and Farhadi}{2018}]{redmon2018yolov3}
Joseph Redmon and Ali Farhadi.
\newblock Yolov3: An incremental improvement.
\newblock {\em arXiv preprint arXiv:1804.02767}, 2018.

\bibitem[\protect\citeauthoryear{Ren \bgroup \em et al.\egroup
  }{2015}]{ren2015faster}
Shaoqing Ren, Kaiming He, Ross Girshick, and Jian Sun.
\newblock Faster r-cnn: Towards real-time object detection with region proposal
  networks.
\newblock {\em Advances in neural information processing systems}, 28, 2015.

\bibitem[\protect\citeauthoryear{Zhao \bgroup \em et al.\egroup
  }{2021}]{zhao2021federated}
Zhongyuan Zhao, Chenyuan Feng, Wei Hong, Jiamo Jiang, Chao Jia, Tony~QS Quek,
  and Mugen Peng.
\newblock Federated learning with non-iid data in wireless networks.
\newblock {\em IEEE Transactions on Wireless Communications}, 21(3):1927--1942,
  2021.

\end{thebibliography}

\end{document}